\documentclass[letterpaper, 10 pt, conference]{ieeeconf}

\usepackage[utf8]{inputenc} %
\usepackage[T1]{fontenc}    %
\usepackage{hyperref}       %
\usepackage{url}            %
\usepackage{booktabs}       %
\usepackage{amsfonts}       %
\usepackage{nicefrac}       %
\usepackage{microtype}      %

\usepackage{times}
\usepackage{epsfig}
\usepackage{graphicx}
\usepackage{amsmath}
\usepackage{amssymb}

\usepackage{subcaption}
\usepackage{verbatim}

\usepackage{balance}

\usepackage{amsmath,amsfonts,bm}

\def\eqref#1{equation~\ref{#1}}

\def\1{\bm{1}}

\def\vtheta{{\bm{\theta}}}
\def\vepsilon{{\bm{\epsilon}}}

\def\vz{{\bm{z}}}

\def\mI{{\bm{I}}}

\def\mO{{\bm{O}}}
\def\mP{{\bm{P}}}

\DeclareMathAlphabet{\mathsfit}{\encodingdefault}{\sfdefault}{m}{sl}
\SetMathAlphabet{\mathsfit}{bold}{\encodingdefault}{\sfdefault}{bx}{n}

\def\emO{{O}}
\def\emP{{P}}

\newcommand{\E}{\mathbb{E}}
\newcommand{\Ls}{\mathcal{L}}
\newcommand{\R}{\mathbb{R}}

\IEEEoverridecommandlockouts                              %

\title{\LARGE \bf
Towards Object Detection from Motion
}

\author{Rico Jonschkowski and Austin Stone%
\thanks{All authors are with Robotics at Google, USA} \thanks{{\tt\small [rjon,austinstone]@google.com}}%
}

\begin{document}

\maketitle
\thispagestyle{empty}
\pagestyle{empty}

\begin{abstract}
We present a novel approach to weakly supervised object detection. Instead of annotated images, our method only requires two short videos to learn to detect a new object: 1) a video of a moving object and 2) one or more ``negative'' videos of the scene without the object. The key idea of our algorithm is to train the object detector to produce physically plausible object motion when applied to the first video and to not detect anything in the second video. With this approach, our method learns to locate objects without any object location annotations. Once the model is trained, it performs object detection on single images. We evaluate our method in three robotics settings that afford learning objects from motion: observing moving objects, watching demonstrations of object manipulation, and physically interacting with objects (see a video summary at \url{https://youtu.be/BH0Hv3zZG_4}).
\end{abstract}

\section{Introduction}

A major bottleneck for object detection in robotics is the need for time-consuming image annotation. We take a step towards overcoming this problem by learning object detection from short videos with minimal supervision. To learn a new object, our approach only requires two short videos, one that shows the object in motion and one that shows the scene without the object. These videos do not need to be synchronized in any way and are therefore easy and fast to generate -- e.g. through human demonstrations or physical interaction of a robot -- which makes this approach very promising for robotics.

The underlying assumption that our method is based on is that \emph{an object is a collection of matter that moves as a unit}. We leverage this fact and use \emph{motion} as a cue for learning object detection. Given a video of a moving object, our approach learns an object detector by optimizing its output to describe physically plausible motion. We additionally collect a \emph{negative} video of the scene without the object and train the object detector to not respond to it, which allows the approach to ignore camera motion and other moving objects. Finally, we use the fact that \emph{objects are spatially local} through a \emph{spatial encoder} architecture that estimates the object's location based on the strongest local activations, which restricts the receptive field and induces robustness to non-local distractions.

\begin{figure}[t]
\centering
\includegraphics[width=0.8\columnwidth]{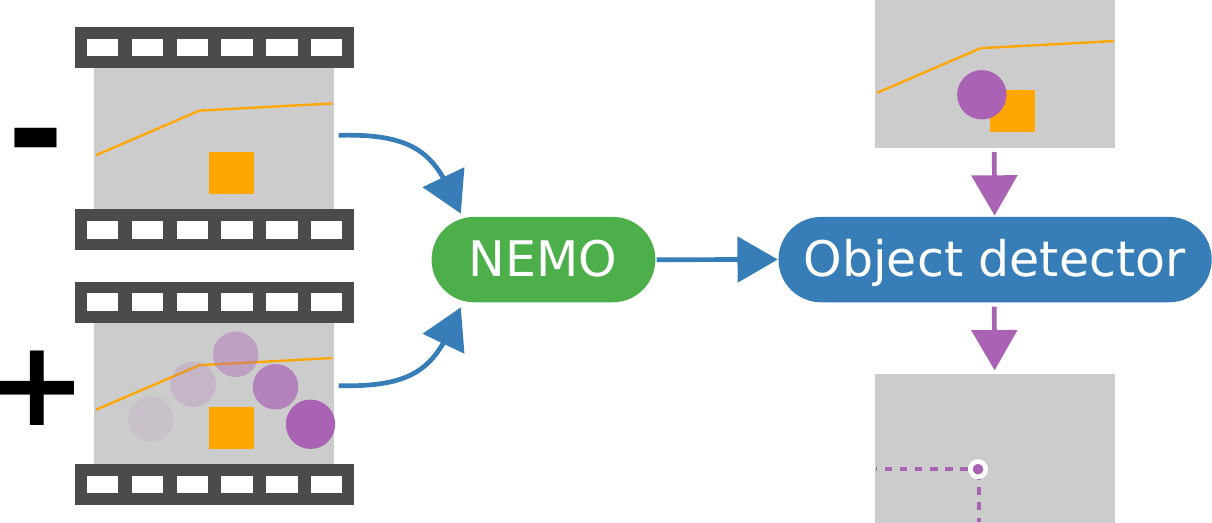}
\caption{Learning to detect an object (purple) from on a video of the object in motion (+) and a video of the scene without the object (-).}
\label{fig:intro}
\end{figure}

Our contribution is a novel approach to weakly supervised learning of object detection, \underline{NEMO}, that uses \underline{n}egative \underline{e}xamples and \underline{mo}tion. NEMO trains a spatial encoder network by optimizing consistency with object motion. It only requires short videos of moving objects that are easy to collect and it does not rely on any pretraining or supervision beyond marking these videos as positive and negative. At inference, the learned model can detect objects regardless of whether they are moving or not because the model works on single images. Note that, although we are evaluating our model on video frames, it does not perform tracking but per frame detection as shown in Figure~\ref{fig:intro}.

\section{Related Work}

This work is related to \emph{physics-based representation learning}, where a latent representation is learned by optimizing consistency with physics. Steward and Ermon~\cite{stewart2017label} learn to map images to latent representations by optimizing consistency with a known dynamics model. Jonschkowski and Brock~\cite{jonschkowski2014learning} and Jonschkowski et al.~\cite{jonschkowski2017pves} make more general assumptions about physical interactions and define them as learning objectives. A number of approaches combine physical assumptions with image reconstruction to learn latent representations~\cite{goroshin2015learning,watter2015embed,finn2016deep}. Gao et al.~\cite{gao2016objectcentric} learn to embed object regions in an image using spatio-temporal consistency. Jang et al.~\cite{jang2018grasp2vec} learn object embeddings from self-supervised interactions based on object permanence. Jayamaran and Grauman~\cite{jayaraman2015learning} learn representations that are equivariant to known ego motion. Sermanet et al.~\cite{sermanet2017time} learn latent representations from multiple synchronous videos of motion. While these approaches are similar to ours in spirit, they learn image embeddings, while we learn to estimate object location in the image. This more constrained object-based representation makes the presented approach particularly robust and efficient.

This paper is also influenced by the idea of \emph{active perception}~\cite{bajcsy1988active}, where action is used to facilitate perception. Motion has been used for a long time to identify and track objects~\cite{lipton1998moving}, to segment them~\cite{fitzpatrick2003first}, to understand their articulation~\cite{katz2008manipulating}, and so on. Recently, this idea has been combined with learning in order to generalize beyond the observed motion, e.g. to learn object segmentation from videos of moving objects~\cite{pathak2017learning} and from videos generated by robot interactions~\cite{pathak2018learning}. This paper goes in the same direction for learning object detection by introducing ideas from representation learning and by leveraging negative examples.

Labeling training videos as positive and negative examples can also be viewed as \emph{weakly supervised learning}, which deals with learning from labels that are only partially informative. Weakly supervised object detection relies on image-wide labels to learn to localize the corresponding objects in the image~\cite{pandey2011scene, oquab2015object}. While these approaches use image-wide labels to replace object location labels, which are more difficult to obtain, this paper goes a step further and only uses per-video labels and compensates this reduction of supervision by adding motion as a cue for learning object detection. Similarly, Tokmakov et al.~\cite{tokmakov2016Motion} learn semantic segmentation by combining weak labels and motion cues from videos, in their case using optical flow, and Hong et al.~\cite{Hong_2017_CVPR} also learn to segment objects use weak semantic labels together with object-ness priors. These works and others in the field of weakly supervised learning leverage large datasets of online videos. Our paper, in contrast, focuses on enabling a robot to learn object detection from its own observations, which means the training data is much more relevant for the immediate tasks of the robot but it also requires our method to be particularly data-efficient.

\section{Object Detection from Negative Examples and Motion (NEMO)}

The key idea of NEMO is to learn how to detect an object from two videos, a \emph{positive video} that shows the target object in motion and a \emph{negative video} of the same scene without that object. These videos are used to optimize two objectives: 1) Learn to detect ``something that moves in a physically plausible way'' in the positive video, such that its location varies over time without having instantaneous jumps, which is defined below as a combination of a \emph{variation loss} and a \emph{slowness loss}. 2) Learn to detect ``something that is present in the positive video but not in the negative video'', which is defined as a \emph{presence loss}. These objectives are used to train a \emph{spatial encoder} network, which estimates the object location based on the strongest activation after a stack of convolutions. Optimization is done by gradient descent. We will now look in detail into each of these components.

\subsection{Network Architecture: Spatial Encoder}

NEMO's network architecture is an extension of the encoder component of a deep spatial autoencoder \cite{finn2016deep} and therefore called a \emph{spatial encoder}. The spatial encoder is a stack of convolutional layers \cite{lecun1998gradient} without striding or pooling. It uses residual connections \cite{he2016deep}, batch normalization \cite{ioffe2015batch}, and ReLU nonlinearities \cite{nair2010rectified}. All experiments in this paper use 8 residual blocks with 32 channels and kernel size 3, which are applied to images scaled to $120\times160$ or $90\times160$. The output has a single channel, followed by a spatial softmax, which produces a probability distribution over the object's location in the image. We obtain a location estimate by taking the mean of that distribution (the spatial softargmax) and estimate the confidence in the network's prediction based on the maximum pre-softmax activation.

Since the main application of this work is to enable learning new objects rapidly and \emph{continually} when they are encountered, we train a separate model for each object in all of our experiments. However, the spatial encoder can be extended to handle $k$ objects by having $k$ channels in the output layer and training those detectors jointly.

Note that our model does not perform any kind of tracking or inference over time because it only takes a single image as input. Video data is only required during training, not during inference.

\begin{figure}
\centering
\includegraphics[width=1.0\columnwidth]{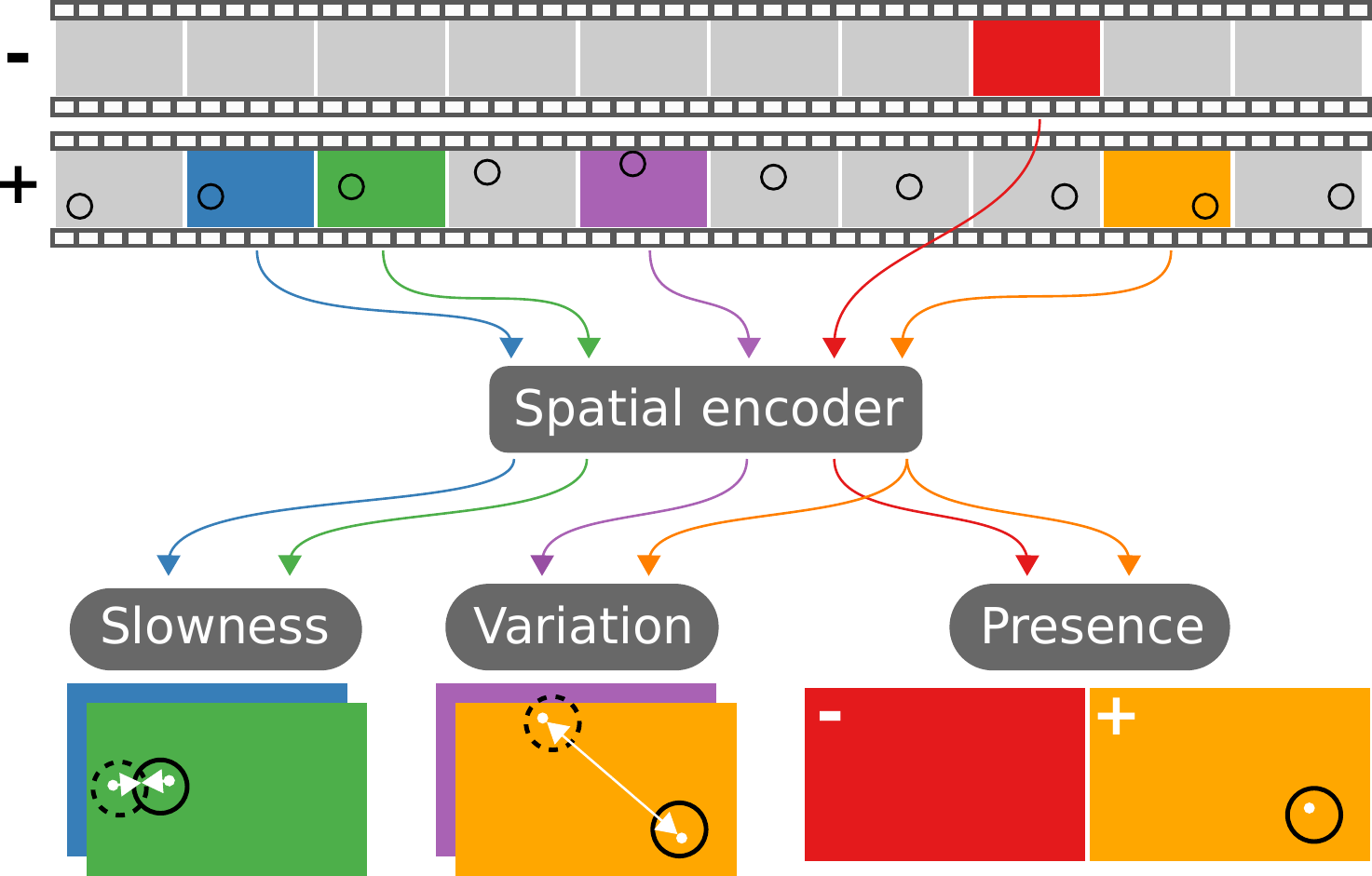}
\caption{{\bf NEMO overview.} Frames from a negative (-) and a positive video (+) with a moving object (black circle) are fed into the spatial encoder. Consecutive frames (blue and green) are optimized for slowness, which produces a gradient that pulls location estimates together. Pairs of distant frames (purple and orange) are optimized for variation, which produces a gradient that pushes location estimates apart. Combinations of positive and negative frames (orange and red) are optimized for detection in the positive frame, the gradient of which increases activations in the positive frame and decreases activations in the negative frame.}
\label{fig:overview}
\end{figure}

\begin{figure*}
\centering
\includegraphics[width=\textwidth]{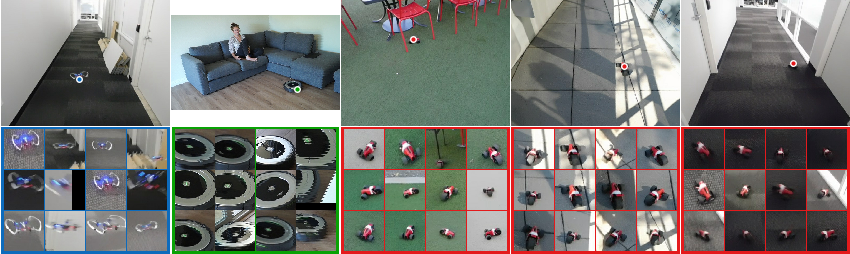}
\caption{{\bf Learning detection by observing moving objects.} \emph{Top:} Five test videos with different scenes and objects. Object detections are shown as colored dots. \emph{Bottom:} Image crops centered at the detected locations in random test frames. \emph{Left to Right:} Four separately trained object detectors for a drone in an office building, a vacuum robot in an apartment, and toy car in two outdoor scenes. The last row shows how the toy car detector generalizes to a new scene.}
\label{fig:movng_q}
\end{figure*}

\begin{figure}
\centering
\includegraphics[width=0.48\columnwidth]{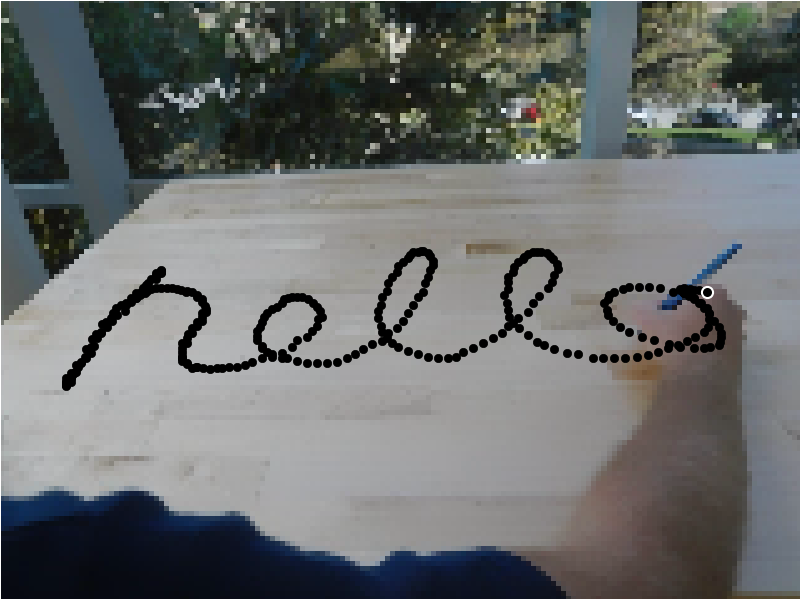} 
\hfill
\includegraphics[width=0.48\columnwidth]{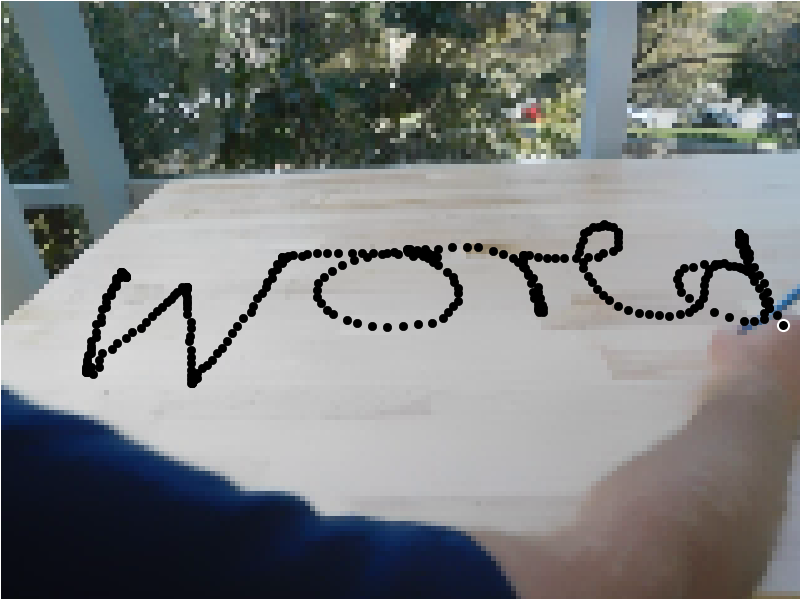}
\caption{{\bf Detecting a pen after demonstration.} Per frame detections in test videos (black dots).
}
\label{fig:pen}
\end{figure}

\subsection{Losses: Variation, Slowness, \& Presence}

The spatial encoder is trained by minimizing a combination of three losses---variation, slowness, and presence (see Figure~\ref{fig:overview}), which are defined here. Let us denote the input image at time $t$ as $\mI^{(t)}\in \R^{h \times w}$ where $h$ and $w$ are the height and width of the image. We will refer to the spatial encoder as $f$ with parameters $\vtheta$, and the output of $f$ before the spatial softmax as $\mO^{(t)}\in \R^{h \times w}$, such that $\mO^{(t)}=f(\mI^{(t)}; \vtheta)$. By applying the spatial softmax across image coordinates $i$ and $j$, we get a probability image $\mP^{(t)}\in \R^{h \times w}$ and its mean $\vz^{(t)}\in\R^2$ normalized to $[-1, 1]$ as
$$\emP^{(t)}_{i,j} = \frac{e^{\emO^{(t)}_{i,j}}}{\sum_{i,j} e^{\emO^{(t)}_{i,j}}} \;\; \text{and} \;\; \vz^{(t)} = \begin{bmatrix} \sum_{i,j} (\frac{2i}{h}-1) \emP^{(t)}_{i,j} \\ \sum_{i,j} (\frac{2j}{w}-1) \emP^{(t)}_{i,j} \end{bmatrix}.$$ 

The first two losses, variation and slowness, operate on the mean $\vz$ in positive frames. Together, they measure whether the detected object location $\vz^{(t)}$ moves in a physically plausible way by comparing pairs of $\vz^{(t)}$ for different $t$.

The \emph{variation loss} encodes the assumption that the target object does not stay still in the video by enforcing that $\vz_{t+d}$ is different from $\vz_{t}$ for $d$ in some range $[d_{\text{min}}, d_{\text{max}}]$. The variation loss measures proximity using $e^{-\text{distance}}$, which is $1$ if $\vz_{t} = \vz_{t+d}$ and goes to $0$ with increasing distance~\cite{jonschkowski2017pves}.
$$\Ls_{\text{variation}}(\vtheta) = \E_{t,d\in[d_{\text{min}}, d_{\text{max}}]} [e^{-\beta||\vz_{t+d}-\vz_{t}||}],$$
where $\beta$ scales how far $\vz_{t}$ and $\vz_{t+d}$ need to be apart and $d_{\text{min}}$ and $d_{\text{max}}$ define for which time differences variation is enforced.
All experiments use $\beta=10$, $d_{\text{min}}=50$, and $d_{\text{max}}=100$.

The \emph{slowness loss} encodes the assumption that objects move with relatively low velocities, i.e., that their locations at time $t$ and $t+1$ are typically close to each other. Consequently, this loss measures the squared distance between $\vz$ in consecutive time steps $t$ and $t+1$, which favors smooth over erratic object trajectories~\cite{wiskott2002slow,jonschkowski2014learning}.
$$\Ls_{\text{slowness}}(\vtheta) = \E_{t} [{||\vz_{t+1}-\vz_{t}||^2}].$$
The \emph{presence loss} encodes the assumption that the object is present in the positive video but not in the negative one. Taking a positive frame $t$ and a negative frame $t^-$, we can compute the probability $q^{(t, t^-)}$ of the object being in the positive frame by computing the spatial softmax jointly over both frames and summing over all pixels. The loss is then defined as negative log probability.
\begin{align*}\Ls_{\text{presence}}(\vtheta) &= \E_{t, t^-} [-\log(q^{(t, t^-)})],\text{where}\\
q^{(t, t^-)} &= \frac{\sum_{i,j} e^{\emO^{(t)}_{i,j}}}{\sum_{i,j} e^{\emO^{(t)}_{i,j}}+e^{\emO^{(t^-)}_{i,j}}}.
\end{align*}

The losses defined above are combined in a weighted sum, $$\Ls(\vtheta) = w_{\text{v}} \Ls_{\text{var.}}(\vtheta) + w_{\text{s}}\Ls_{\text{slown.}}(\vtheta) + w_{\text{p}}\Ls_{\text{pres.}}(\vtheta),$$ where the weights were chosen such that all gradients have the same order of magnitude. All experiments use $w_{\text{v}} = 2$, $w_{\text{s}} = 10$, and $w_{\text{p}} = 1$. 

\subsection{Optimization}

The losses are optimized from minibatches of size $b$, such that each minibatch includes $b$ samples of consecutive frames $\{(\mI^{(t)}, \mI^{(t+1)})\}_t$ and $b$ samples of frames $d\in[d_{\text{min}}, d_{\text{max}}]$ steps apart $\{(\mI^{(t)}, \mI^{(t+d)})\}_{t,d}$, which are used to compute the variation and slowness losses. The presence loss uses all combinations of the positive frames in $\{(\mI^{(t)}, \mI^{(t+d)})\}_{t,d}$ with $b$ negative frames $\{\mI^{(t^{-})}\}_{t^{-}}$ resulting in $2b^2$ pairs to average over. All experiments use $b=10$. For numerical stability of the gradient computation, Gaussian noise $\vepsilon \sim \mathcal{N} (\mu=0 , \sigma=10^{-5})$ is added to $\vz_{t}$. The loss $\Ls(\vtheta)$ is optimized using the adaptive gradient descent method Adam~\cite{kingma2015adam} with default parameters and $m=50$ random restarts. Our implementation is based on TensorFlow~\cite{tensorflow} and Keras~\cite{keras}.

\begin{figure*}
\centering
\includegraphics[width=\textwidth]{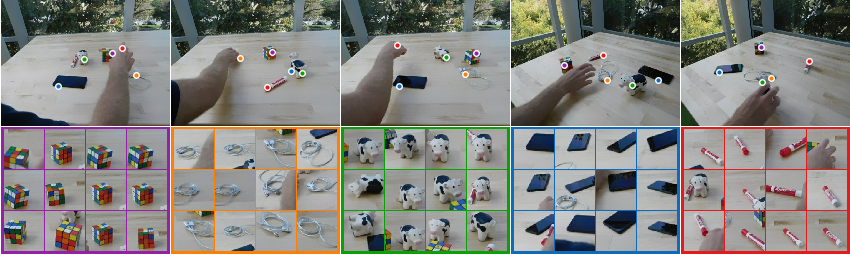}
\caption{{\bf Learning multiple objects from demonstration:} a USB cable (orange), a toy cow (green), a toy cube (purple), a marker (red), and a phone (blue). \emph{Top:} Random frames in a multi-object video after training on single object videos. \emph{Bottom:} Image crops centered at the detected object locations in random test frames. \emph{Right:} Generalization to a different viewpoint.}
\label{fig:demon_q}
\end{figure*}

\begin{figure*}
\begin{center}
\includegraphics[width=\linewidth]{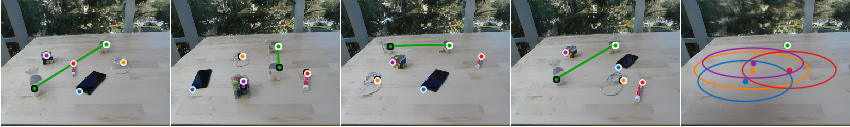}
\includegraphics[width=\linewidth]{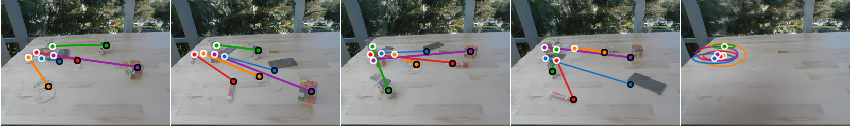}
\caption{{\bf Demonstrating manipulation tasks with learned object representations.} \emph{Left four rows:} Four demonstrations of a task with overlayed first and the last frames. First frame object detections (black circles) are connected to last frame detections (white circles). \emph{Right:} Mean and three standard deviations of last frame detections across all demonstrations reveal the goal of the tasks; white circle indicates task-relevance based on if average object motion is at least 10 pixels. \emph{Top to bottom:} Two manipulation tasks: 1) Move the cow (green) to top center. 2) Move all objects to the top left.}
\label{fig:demo}
\end{center}
\end{figure*}

\section{Experiments}
\label{sec:exp}

\begin{figure*}
\centering
\includegraphics[width=\textwidth]{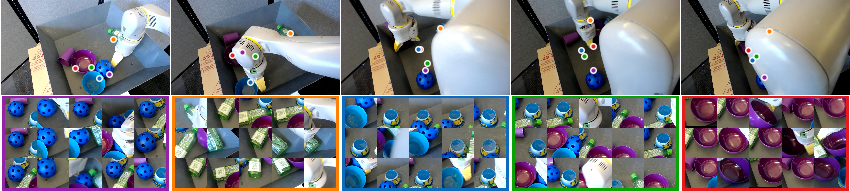}
\caption{{\bf Learning object detection from interaction.} \emph{Top:} Image sequence from a multi-object test video after training on single object interactions. \emph{Bottom:} Image crops centered at the detected object locations. Our method learned objects 1 (blue ball), 2 (green bottle), 3 (blue bowl), and 5 (purple bowl), but failed to learn object 4 (purple cup).}
\label{fig:interaction_qualitative}
\end{figure*}

We evaluated NEMO in three settings that afford object detection from motion: 
\begin{enumerate}
    \item Learning to detect moving objects by observing them (see Figure~\ref{fig:movng_q})
    \item Learning to detect static objects from human demonstrations (see Figures~\ref{fig:pen} and~\ref{fig:demon_q})
    \item Learning to detect static objects by physically interacting with them (see Figure~\ref{fig:interaction_qualitative})
\end{enumerate}  
In all settings, our method was trained on short (less than five minutes) positive and negative videos and then tested on individual frames from another video. Note that NEMO does not perform tracking. All results show per frame detection.

Sections~\ref{sec:setting1} to \ref{sec:setting3} demonstrate the versatility and wide applicability of our method. A comparison to across all three settings in Section~\ref{sec:comparison} quantifies the accuracy of NEMO relative to existing methods.
The key results of our experiments are: a) Our method is able to discover objects without any image level annotations from a few short videos of moving objects. b) Our method is robust to distracting motion of the camera, the arm, and other moving objects as well as to substantial occlusions during training and testing. c) In a comparison to other methods, NEMO outperforms supervised object detection trained on COCO~\cite{lin2014microsoft},
template matching, and tracking methods, even though it is the only method among these that does not require any image annotations. A video summarizing the results can be found here: \url{https://youtu.be/BH0Hv3zZG_4}

\subsection{Learning from Observation}
\label{sec:setting1}

The first setting that we evaluate is about learning to detect moving objects (such as vehicles, humans, other robots, etc.) from observing them with a moving camera. The main difficulty of object detection in this scenario is disambiguating object motion and camera motion.

Each object detector was learned from a positive and a negative video of about two minutes and evaluated in frames of a test video. The results show that NEMO can discover moving objects despite camera motion for a variety of different scenes and objects. Figure~\ref{fig:movng_q} shows representative object detections for different moving objects. The quantitative evaluation for this setting can be found in Figure~\ref{fig:movng_c} and will be discussed in Section~\ref{sec:comparison}.

\begin{figure}
\centering
\includegraphics[width=0.48\columnwidth]{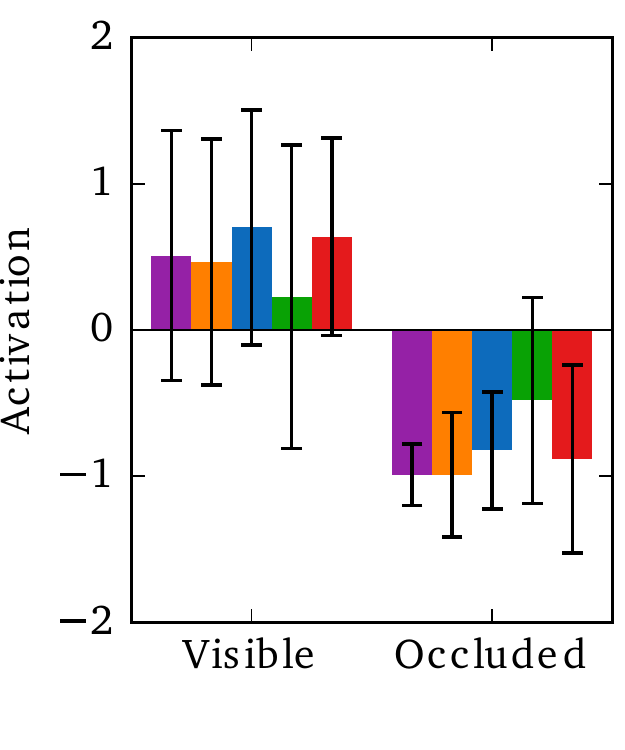}
\hfill
\includegraphics[width=0.48\columnwidth]{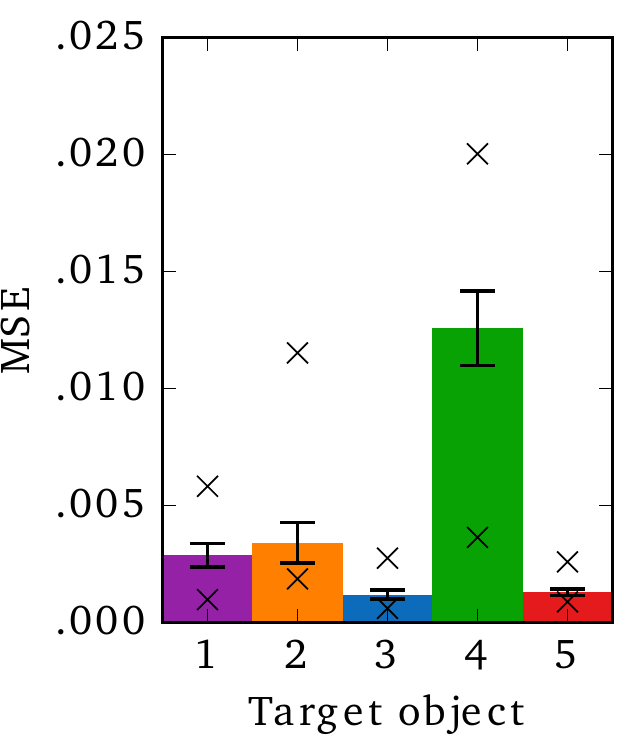}
\caption{{\bf Detailed results for learning from interaction.} \emph{Left:}  Maximum pre-softmax activation in the encoder is a good estimate for whether the object is present in the scene (error bars show standard deviations). \emph{Right:} Results per object with standard errors and minimum and maximum show that the method has the most difficulty learning object 4 (purple cup).}
\label{fig:int_d}
\end{figure}

\subsection{Learning from Demonstration}
\label{sec:setting2}

Most objects do not move on their own, but objects relevant for robot manipulation typically can \emph{be moved} in order to generate the necessary motion cues for learning them. To illustrate this idea, we had a human move a pen in spirals across a table to provide a positive example and move their hand without the pen for a negative example, each for about one minute. We applied our method on these videos and tested the trained object detector on new videos of writing ``hello" and ``world''. The results show per frame object detections that are very stable over time and thereby allow NEMO to accurately trace the objects trajectory (see Figure~\ref{fig:pen}).

\begin{figure*}
\begin{subfigure}[b]{.352\linewidth}
\includegraphics[width=\linewidth]{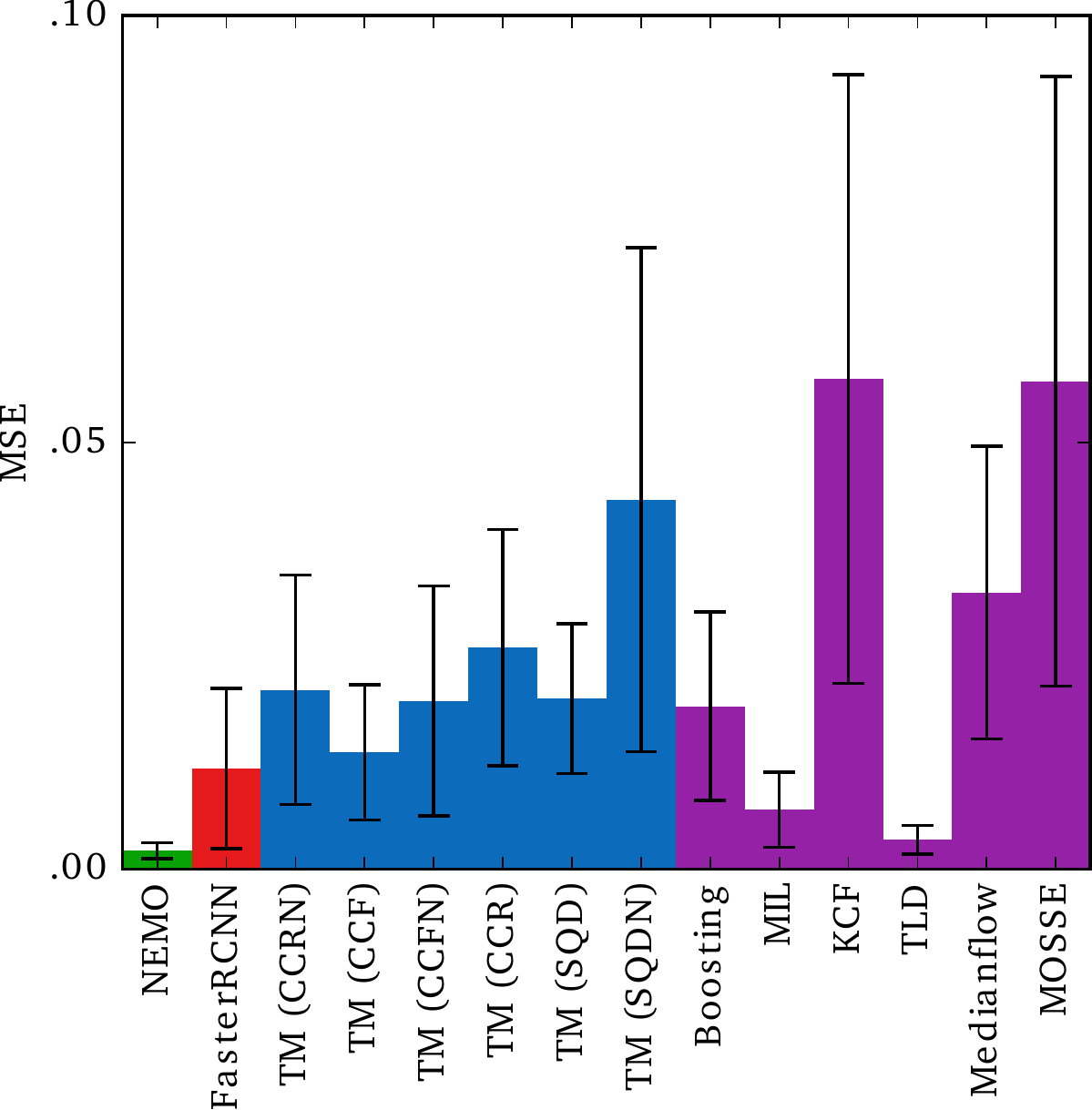}
\caption{Learning from observation}
\label{fig:movng_c}
\end{subfigure}
\hfill
\begin{subfigure}[b]{.315\linewidth}
\includegraphics[width=\linewidth]{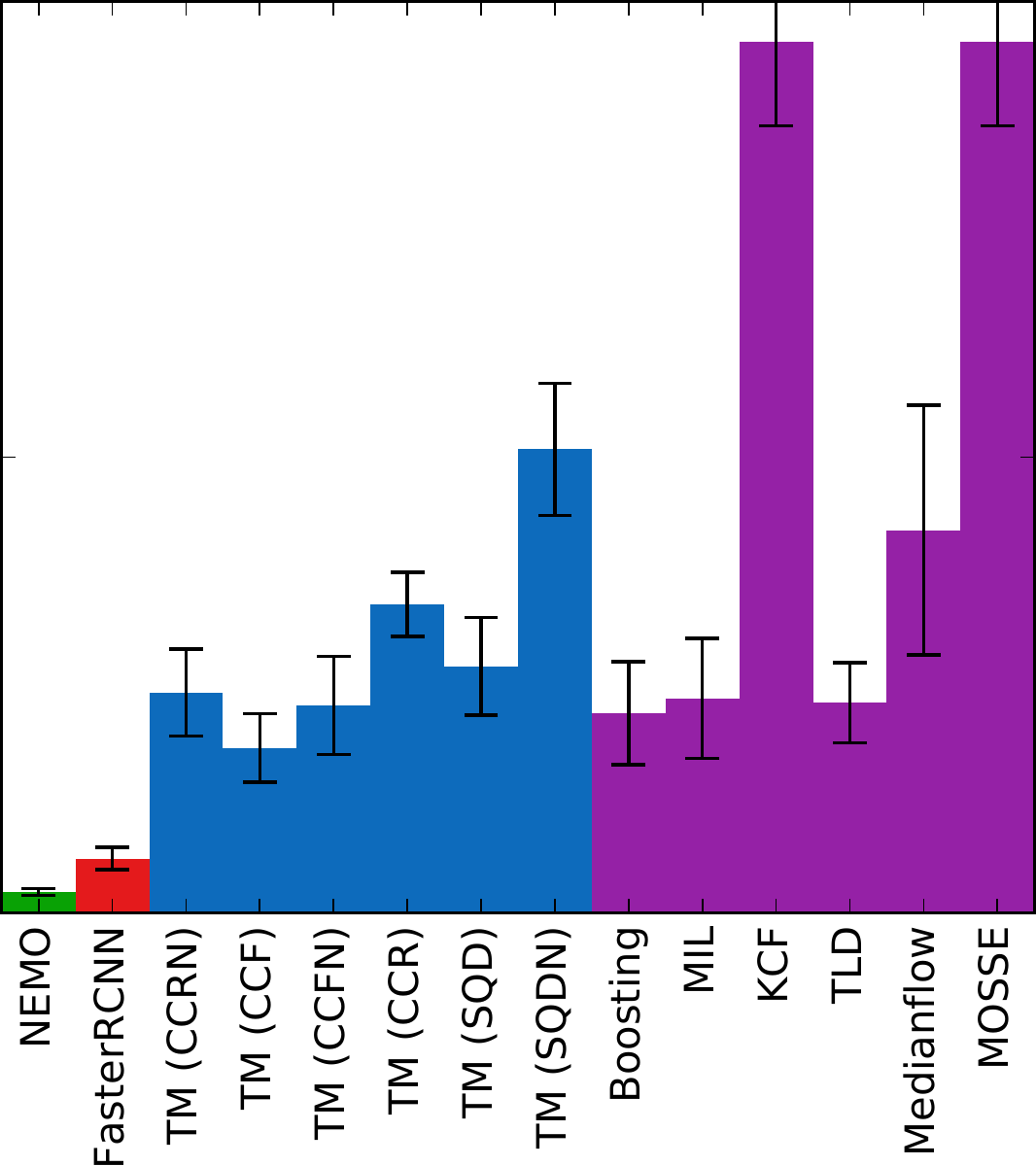}
\caption{Learning from demonstration}
\label{fig:demon_c}
\end{subfigure}
\hfill
\begin{subfigure}[b]{.315\linewidth}
\includegraphics[width=\linewidth]{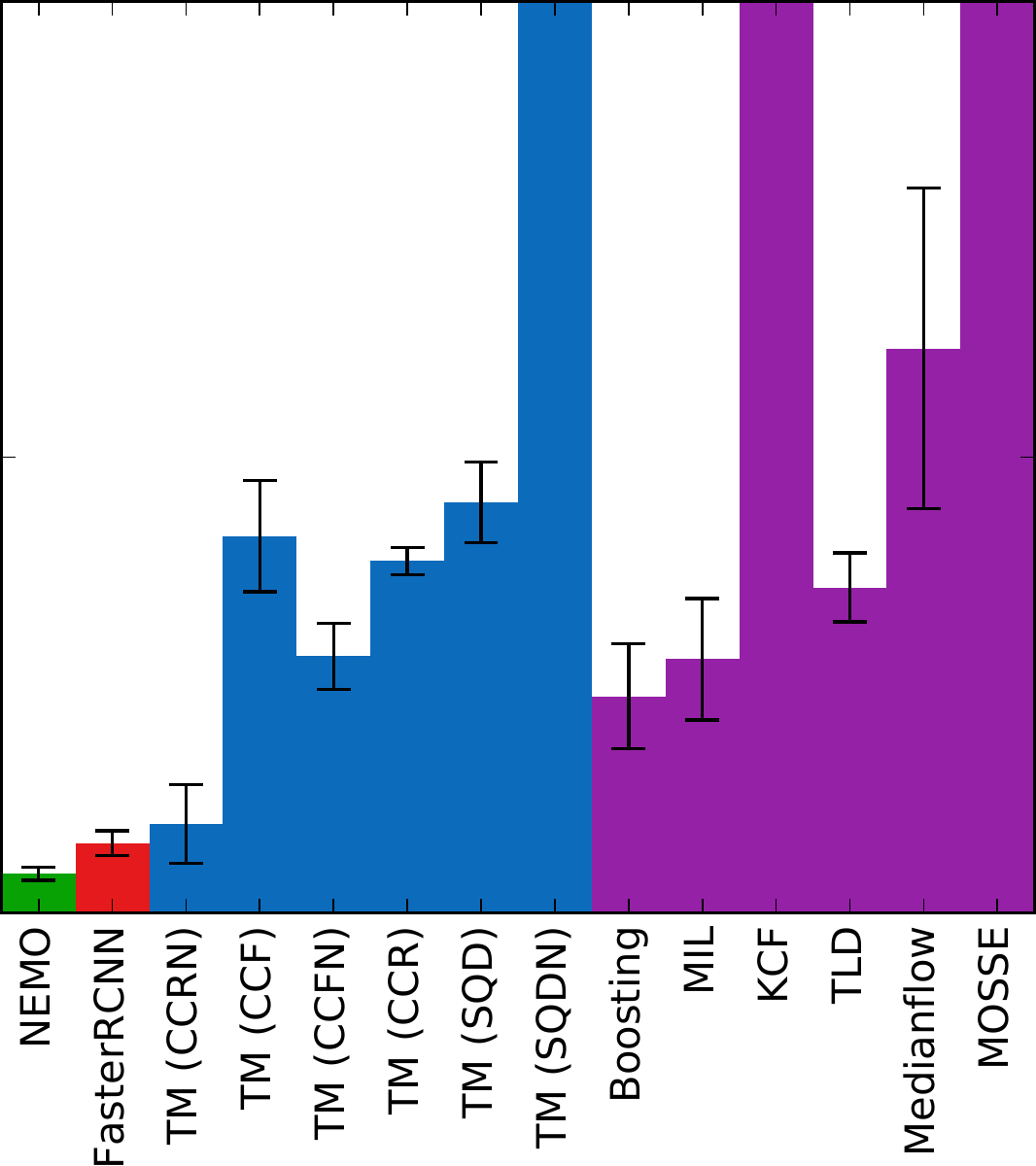}
\caption{Learning from interaction}
\label{fig:int_c}
\end{subfigure}
\caption{Results for different methods for the three settings described above. Error bars show standard errors. Template matching methods are shown in blue; tracking methods are shown in purple. NEMO consistently achieved the lowest error.}
\label{fig:comparison}
\end{figure*}

The setting of having a human move objects combines well with learning from demonstration in robotics, where a teacher shows a new task to a robot. Before demonstrating the task, the teacher can move the relevant objects to train object detection. The learned objects can then serve as a high level representation for interpreting the demonstration and extracting the goal of the task.
We investigate this scenario by having a human move different objects on a table. More specifically, the training data for this task are five videos each about two minutes long of five different objects moved across a table. We use this data to train five object detectors, one per object. For each detector, the positive example is one of the object videos and the negative example is a concatenation of the four other videos not containing the object in the positive video. These resulting detectors are tested in a video in which all five objects are being randomly rearranged on the table (see Figure~\ref{fig:demon_q}). The results show correct detections even under heavy occlusions by the hand and other objects and generalization to a different viewpoint. We quantitatively evaluate this setting in Figure~\ref{fig:demon_c} and Section~\ref{sec:comparison}.

After training the object detectors as described, we investigate if they could be used for learning from demonstration. By applying the object detectors to demonstrations of pick-and-place tasks and comparing object locations in the first and last frame of a demonstration, we can find out which objects were moved to which locations. Simple statistics over the final object locations across all demonstrations of a task describe the task in a way that could be used as the goal for a planning algorithm or as reward for reinforcement learning (see Figure~\ref{fig:demo}).

\subsection{Learning from Interaction}
\label{sec:setting3}

In this setting, we test whether our method can learn to detect objects from a robot interacting with them. Ultimately, we envision a robot learning about the objects around it testing hypotheses about potential objects by pushing and pulling on things in the environment. For now, we have a robot execute a simple open-loop trajectory that moves its arm in a circular sweeping motion through a bin containing objects. What makes this setting more challenging than using human demonstrations is that the objects move less and are occluded by the robot arm for significant periods in the training videos.

The training data was collected by putting one object at a time into a bin and having the robot interact with it for five minutes. Based on these videos, NEMO was applied to learn an object detector for each object using the other videos as negative examples. The learned object detectors were tested in a setting where all objects were placed in the bin and again randomly moved by the robot (see Figure~\ref{fig:interaction_qualitative}). In the bottom of the figure, from left to right, we can see that the method learned to track the ball (purple), the bottle (orange), the small bowl (blue), that it failed to learn the purple cup (green), and that it successfully learned the large bowl (red). A comparison to other methods is shown in Figure~\ref{fig:int_c}.

Figure~\ref{fig:int_d} shows that the activation of the object detector can be used to estimate whether an object is currently visible or not. It also confirms that our method indeed struggled with object 4, the purple cup. Inspection of the training video showed that the robot gripper often moved over the purple cup instead of pushing it, which apparently did not produce enough motion to learn the object. The other four objects were learned despite substantial occlusions during training, which shows the feasibility of learning object detection from interaction using NEMO. For a video of the interaction, see \url{https://youtu.be/BH0Hv3zZG_4?t=213}.

\subsection{Comparison to Other Methods}
\label{sec:comparison}

Comparison to other methods is complicated by the fact that most other techniques require more supervision than NEMO and have not been designed to leverage the object information latent in positive and negative video examples. Therefore, to perform a fair quantitative comparison, we selected other related methods and gave them the additional supervision that they need in order to apply them to our data.

We include a range of template matching and tracking methods in our comparison because they need relatively little supervision. For those methods, we provided a bounding box of the target object in the first frame of the test video. The exact methods we are comparing to are template matching using normalized and unnormalized cross-correlation, cross-coefficient, and squared distance scores ~\cite{lewis1995fast} and the following tracking methods--- OLB~\cite{grabner2006real}, MIL~\cite{babenko2009visual}, KCF~\cite{henriques2015high}, TLD~\cite{kalal2012tracking}, Medianflow~\cite{kalal2010forward}, MOSSE~\cite{bolme2010visual}---using their OpenCV~\cite{opencv_library} implementations. 

We also compare to supervised object detection in the form of the region proposal network FasterRCNN~\cite{ren2015faster} trained on the COCO dataset~\cite{lin2014microsoft}. Following other approaches in the literature~\cite{pirk2019online}, we use this model as an ``objectness detector'' and disregard the predicted object class. For computing the error of this model, we select the nearest predicted bounding box to the target object, allowing the same bounding box to count as prediction for multiple objects.

We compare our results in all of the above settings to these methods. For our comparison, we labeled object positions in 50 frames in each test video for all three settings and compare the methods based on mean squared error (MSE) normalized by the image diagonal. Figure \ref{fig:comparison} shows that NEMO (green) performs well in this comparison. While TLD and MIL tracking (purple) work well in the first setting, they fail in the other settings due to phases where the object gets occluded. FasterRCNN (red) works better in those settings but has a higher error in the first setting. Other tracking and template matching methods perform significantly worse. NEMO consistently achieves low errors across all three settings. But more importantly, it does so despite being the only method that does not require any image annotations. These results show the advantage of adapting to the given set of objects using weakly supervised learning and hint at the potential of future work on object detection from motion.

\section{Conclusion}

We have proposed a novel approach for learning object detection that is simple, data-efficient, and requires minimal supervision. We have demonstrated the versatility of this approach that could enable robots to learn object detection without image level annotations from only a few minutes of video. Future work could improve transfer to new scenes, learn from a continuous video stream, and produce more expressive representations such as bounding boxes, object segments, key points, or poses. We believe that this is a promising research direction that could ultimately facilitate robotic manipulation by enabling robots to rapidly learn new objects with minimal supervision.

\section*{Acknowledgments}
\small{We would like to thank Mohi Khansari for conducting the robot experiment and Anelia Angelova, Kurt Konolige, Tomas Pfister, Alexander Toshev, Chelsea Finn, Pierre Sermanet, Tsung-Yi Lin, and Vincent Vanhoucke for their constructive feedback on this work.}

\clearpage

\balance
\bibliography{paper}

\begin{thebibliography}{10}

\bibitem{stewart2017label}
R.~Stewart and S.~Ermon, ``Label-free supervision of neural networks with
  physics and domain knowledge.,'' in {\em {AAAI Conference on Artificial
  Intelligence}}, pp.~1--7, 2017.

\bibitem{jonschkowski2014learning}
R.~Jonschkowski and O.~Brock, ``Learning state representations with robotic
  priors,'' {\em Autonomous Robots}, vol.~39, no.~3, pp.~407--428, 2015.

\bibitem{jonschkowski2017pves}
R.~Jonschkowski, R.~Hafner, J.~Scholz, and M.~Riedmiller, ``Pves:
  Position-velocity encoders for unsupervised learning of structured state
  representations,'' in {\em {New Frontiers for Deep Learning in Robotics
  Workshop at RSS}}, 2017.

\bibitem{goroshin2015learning}
R.~Goroshin, M.~F. Mathieu, and Y.~LeCun, ``Learning to linearize under
  uncertainty,'' in {\em {Advances in Neural Information Processing Systems
  (NIPS)}}, pp.~1234--1242, 2015.

\bibitem{watter2015embed}
M.~Watter, J.~Springenberg, J.~Boedecker, and M.~Riedmiller, ``Embed to
  control: A locally linear latent dynamics model for control from raw
  images,'' in {\em {Advances in Neural Information Processing Systems
  (NIPS)}}, pp.~2746--2754, 2015.

\bibitem{finn2016deep}
C.~Finn, X.~Y. Tan, Y.~Duan, T.~Darrell, S.~Levine, and P.~Abbeel, ``Deep
  spatial autoencoders for visuomotor learning,'' in {\em {IEEE International
  Conference on Robotics and Automation (ICRA)}}, pp.~512--519, 2016.

\bibitem{gao2016objectcentric}
R.~Gao, D.~Jayaraman, and K.~Grauman, ``Object-centric representation learning
  from unlabeled videos,'' in {\em {Asian Conference on Computer Vision
  (ACCV)}}, November 2016.

\bibitem{jang2018grasp2vec}
E.~Jang, C.~Devin, V.~Vanhoucke, and S.~Levine, ``Grasp2vec: Learning object
  representations from self-supervised grasping,'' {\em Proceedings of Machine
  Learning Research}, 2018.

\bibitem{jayaraman2015learning}
D.~Jayaraman and K.~Grauman, ``Learning image representations tied to
  ego-motion,'' in {\em {IEEE International Conference on Computer Vision
  (ICCV)}}, pp.~1413--1421, 2015.

\bibitem{sermanet2017time}
P.~Sermanet, C.~Lynch, Y.~Chebotar, J.~Hsu, E.~Jang, S.~Schaal, and S.~Levine,
  ``Time-contrastive networks: Self-supervised learning from video,'' {\em
  arXiv preprint arXiv:1704.06888}, 2017.

\bibitem{bajcsy1988active}
R.~Bajcsy, ``Active perception,'' in {\em IEEE Proceedings}, vol.~76,
  pp.~996--1006, 1988.

\bibitem{lipton1998moving}
A.~J. Lipton, H.~Fujiyoshi, and R.~S. Patil, ``Moving target classification and
  tracking from real-time video,'' in {\em Fourth IEEE Workshop on Applications
  of Computer Vision (WACV)}, pp.~8--14, IEEE, 1998.

\bibitem{fitzpatrick2003first}
P.~Fitzpatrick, ``First contact: an active vision approach to segmentation,''
  in {\em IEEE/RSJ International Conference on Intelligent Robots and Systems
  (IROS)}, vol.~3, pp.~2161--2166, IEEE, 2003.

\bibitem{katz2008manipulating}
D.~Katz and O.~Brock, ``Manipulating articulated objects with interactive
  perception,'' in {\em IEEE International Conference on Robotics and
  Automation (ICRA)}, pp.~272--277, IEEE, 2008.

\bibitem{pathak2017learning}
D.~Pathak, R.~B. Girshick, P.~Doll{\'a}r, T.~Darrell, and B.~Hariharan,
  ``Learning features by watching objects move.,'' in {\em {IEEE Conference on
  Computer Vision and Pattern Recognition (CVPR)}}, vol.~1, p.~7, 2017.

\bibitem{pathak2018learning}
D.~Pathak, Y.~Shentu, D.~Chen, P.~Agrawal, T.~Darrell, S.~Levine, and J.~Malik,
  ``Learning instance segmentation by interaction,'' in {\em Proceedings of the
  IEEE Conference on Computer Vision and Pattern Recognition Workshops},
  pp.~2042--2045, 2018.

\bibitem{pandey2011scene}
M.~Pandey and S.~Lazebnik, ``Scene recognition and weakly supervised object
  localization with deformable part-based models,'' in {\em International
  Conference on Computer Vision (ICCV)}, pp.~1307--1314, IEEE Computer Society,
  2011.

\bibitem{oquab2015object}
M.~Oquab, L.~Bottou, I.~Laptev, and J.~Sivic, ``Is object localization for
  free?-weakly-supervised learning with convolutional neural networks,'' in
  {\em {IEEE Conference on Computer Vision and Pattern Recognition (CVPR)}},
  pp.~685--694, 2015.

\bibitem{tokmakov2016Motion}
C.~S. Pavel~Tokmakov, Karteek~Alahari, ``Weakly-supervised semantic
  segmentation using motion cues,'' in {\em {European Conference on Computer
  Vision (ECCV)}}, pp.~388--404, 2016.

\bibitem{Hong_2017_CVPR}
S.~Hong, D.~Yeo, S.~Kwak, H.~Lee, and B.~Han, ``Weakly supervised semantic
  segmentation using web-crawled videos,'' in {\em The IEEE Conference on
  Computer Vision and Pattern Recognition (CVPR)}, July 2017.

\bibitem{lecun1998gradient}
Y.~LeCun, L.~Bottou, Y.~Bengio, and P.~Haffner, ``Gradient-based learning
  applied to document recognition,'' {\em Proceedings of the IEEE}, vol.~86,
  no.~11, pp.~2278--2324, 1998.

\bibitem{he2016deep}
K.~He, X.~Zhang, S.~Ren, and J.~Sun, ``Deep residual learning for image
  recognition,'' in {\em {IEEE Conference on Computer Vision and Pattern
  Recognition (CVPR)}}, pp.~770--778, 2016.

\bibitem{ioffe2015batch}
S.~Ioffe and C.~Szegedy, ``Batch normalization: Accelerating deep network
  training by reducing internal covariate shift,'' in {\em {International
  Conference on Machine Learning (ICML)}}, pp.~448--456, 2015.

\bibitem{nair2010rectified}
V.~Nair and G.~E. Hinton, ``Rectified linear units improve restricted boltzmann
  machines,'' in {\em {International Conference on Machine Learning (ICML)}},
  pp.~807--814, 2010.

\bibitem{wiskott2002slow}
L.~Wiskott and T.~J. Sejnowski, ``Slow feature analysis: Unsupervised learning
  of invariances,'' {\em {Neural Computation}}, vol.~14, no.~4, pp.~715--770,
  2002.

\bibitem{kingma2015adam}
D.~P. Kingma and J.~L. Ba, ``Adam: Amethod for stochastic optimization,'' in
  {\em {International Conference on Learning Representations (ICLR)}}, 2015.

\bibitem{tensorflow}
M.~Abadi, A.~Agarwal, P.~Barham, E.~Brevdo, Z.~Chen, C.~Citro, G.~S. Corrado,
  A.~Davis, J.~Dean, M.~Devin, S.~Ghemawat, I.~Goodfellow, A.~Harp, G.~Irving,
  M.~Isard, Y.~Jia, R.~Jozefowicz, L.~Kaiser, M.~Kudlur, J.~Levenberg,
  D.~Man\'{e}, R.~Monga, S.~Moore, D.~Murray, C.~Olah, M.~Schuster, J.~Shlens,
  B.~Steiner, I.~Sutskever, K.~Talwar, P.~Tucker, V.~Vanhoucke, V.~Vasudevan,
  F.~Vi\'{e}gas, O.~Vinyals, P.~Warden, M.~Wattenberg, M.~Wicke, Y.~Yu, and
  X.~Zheng, ``{TensorFlow}: Large-scale machine learning on heterogeneous
  systems,'' 2015.
\newblock Software available from tensorflow.org.

\bibitem{keras}
F.~Chollet {\em et~al.}, ``Keras.'' \url{https://keras.io}, 2015.

\bibitem{lin2014microsoft}
T.-Y. Lin, M.~Maire, S.~Belongie, J.~Hays, P.~Perona, D.~Ramanan,
  P.~Doll{\'a}r, and C.~L. Zitnick, ``Microsoft coco: Common objects in
  context,'' in {\em {European Conference on Computer Vision (ECCV)}},
  pp.~740--755, Springer, 2014.

\bibitem{lewis1995fast}
J.~P. Lewis, ``Fast template matching,'' in {\em Vision interface}, vol.~95,
  pp.~15--19, 1995.

\bibitem{grabner2006real}
H.~Grabner, M.~Grabner, and H.~Bischof, ``Real-time tracking via on-line
  boosting,'' in {\em {British Machine Vision Conference (BMVC)}}, vol.~1,
  p.~6, 2006.

\bibitem{babenko2009visual}
B.~Babenko, M.-H. Yang, and S.~Belongie, ``Visual tracking with online multiple
  instance learning,'' in {\em {IEEE Conference on Computer Vision and Pattern
  Recognition (CVPR)}}, pp.~983--990, 2009.

\bibitem{henriques2015high}
J.~F. Henriques, R.~Caseiro, P.~Martins, and J.~Batista, ``High-speed tracking
  with kernelized correlation filters,'' {\em IEEE Transactions on Pattern
  Analysis and Machine Intelligence}, vol.~37, no.~3, pp.~583--596, 2015.

\bibitem{kalal2012tracking}
Z.~Kalal, K.~Mikolajczyk, J.~Matas, {\em et~al.},
  ``Tracking-learning-detection,'' {\em IEEE Transactions on Pattern Analysis
  and Machine Intelligence}, vol.~34, no.~7, p.~1409, 2012.

\bibitem{kalal2010forward}
Z.~Kalal, K.~Mikolajczyk, and J.~Matas, ``Forward-backward error: Automatic
  detection of tracking failures,'' in {\em {International Conference on
  Pattern Recognition (ICPR)}}, pp.~2756--2759, 2010.

\bibitem{bolme2010visual}
D.~S. Bolme, J.~R. Beveridge, B.~A. Draper, and Y.~M. Lui, ``Visual object
  tracking using adaptive correlation filters,'' in {\em {IEEE Conference on
  Computer Vision and Pattern Recognition (CVPR)}}, pp.~2544--2550, 2010.

\bibitem{opencv_library}
G.~Bradski, ``{The OpenCV Library},'' {\em Dr. Dobb's Journal of Software
  Tools}, 2000.

\bibitem{ren2015faster}
S.~Ren, K.~He, R.~Girshick, and J.~Sun, ``Faster r-cnn: Towards real-time
  object detection with region proposal networks,'' in {\em {Advances in Neural
  Information Processing Systems (NIPS)}}, pp.~91--99, 2015.

\bibitem{pirk2019online}
S.~Pirk, M.~Khansari, Y.~Bai, C.~Lynch, and P.~Sermanet, ``Online object
  representations with contrastive learning,'' {\em arXiv preprint
  arXiv:1906.04312}, 2019.

\end{thebibliography}
\bibliographystyle{ieeetr}

\end{document}